\documentclass{article}
\usepackage{ijcai25}

\usepackage{times}
\usepackage{soul}
\usepackage{url}
\usepackage[hidelinks]{hyperref}
\usepackage[utf8]{inputenc}
\usepackage[small]{caption}
\usepackage{graphicx}
\usepackage{amsmath}
\usepackage{amsthm}
\usepackage{booktabs}
\usepackage{algorithm}
\usepackage{algorithmic}
\usepackage[switch]{lineno}

\usepackage{newfloat}
\usepackage{listings}
\usepackage[edges]{forest}
\usepackage{multirow}

\usepackage{makecell}
\usepackage{color}
\definecolor{span_pink}{RGB}{255,122,179}
\definecolor{entity_blue}{RGB}{148,169,216}
\definecolor{lightcoral}{rgb}{0.94, 0.5, 0.5}
\definecolor{lightgreen}{rgb}{0.56, 0.93, 0.56}
\definecolor{capri}{rgb}{0.0, 0.75, 1.0}
\definecolor{carminepink}{rgb}{0.92, 0.3, 0.26}

\definecolor{hidden-draw}{RGB}{205, 44, 36}
\definecolor{hidden-blue}{RGB}{194,232,247}
\definecolor{hidden-orange}{RGB}{243,202,120}
\definecolor{hidden-yellow}{RGB}{242,244,193}
\definecolor{tree-level-1}{RGB}{245,20,85}
\definecolor{tree-level-2}{RGB}{246,86,118}
\definecolor{tree-level-3}{RGB}{248,177,193}
\definecolor{tree-leaf}{RGB}{176,230,198}

\definecolor{Self}{RGB}{255,0,128}
\definecolor{Ensemble}{RGB}{0,127,255}
\definecolor{Iterative}{RGB}{153,51,255}

\definecolor{exemplar1}{RGB}{136,98,148}
\definecolor{exemplar2}{RGB}{148,210,242}
\definecolor{knowledge1}{RGB}{249,219,152}
\definecolor{knowledge2}{RGB}{255,245,220}
\usepackage{xcolor}
\usepackage{xcolor,colortbl}
\definecolor{myorange}{RGB}{251,246,237}
\definecolor{mypurple}{RGB}{242,237,243}
\definecolor{myblue}{RGB}{211,237,242}

\definecolor{mygraym}{RGB}{206,203,203}
\definecolor{mygrayp}{RGB}{206,203,203}
\definecolor{mygrayb}{RGB}{206,203,203}

\newcommand{\lcone}{300}
\newcommand{\lctwo}{200}
\newcommand{\lcthree}{120}

\newcommand{\lcfive}{70}

\tikzstyle{my-box}=[
    rectangle,
    draw=hidden-draw!50,
    rounded corners,
    text opacity=1,
    minimum height=1.0em,
    minimum width=28em,
    inner sep=2pt,
    align=left,
    fill opacity=.5,
]
\tikzstyle{prevent_leaf}=[my-box,
    fill=mypurple!90, text=black, align=left,font=\scriptsize,
    inner xsep=2pt,
    inner ysep=4pt,
]
\tikzstyle{detect_leaf}=[my-box, 
    fill=myblue!\lcfive, text=black, align=left,font=\scriptsize,
    inner xsep=2pt,
    inner ysep=4pt,
]
\tikzstyle{misinfo_leaf}=[my-box,
    fill=myorange!\lcfive, text=black, align=left,font=\scriptsize,
    inner xsep=2pt,
    inner ysep=4pt,
]

\title{A Survey of fMRI to Image  Reconstruction}

\author{
Weiyu Guo$^1$
\and
Guoying Sun$^2$\and
Jianxiang He$^1$\and
Tong Shao$^2$\and
Shaoguang Wang$^1$\and
Ziyang Chen$^1$\and
Meisheng Hong$^3$\and
Ying Sun$^1$\And
Hui Xiong$^1$
\affiliations
$^1$Thrust of Artificial Intelligence, The Hong Kong University of Science and Technology (Guangzhou)
\\
$^2$College of Computer Science and Technology, Harbin Institute of Technology (Shenzhen)\\
$^3$School of Control Science and Engineering, Shandong University\\
\emails
\{wguo395\}@connect.hkust-gz.edu.cn,
\{yings, xionghui\}@ust.hk
}

\begin{document}

\maketitle

\begin{abstract}
Functional magnetic resonance imaging (fMRI)-based image reconstruction plays a pivotal role in decoding human perception, with applications in neuroscience and brain-computer interfaces. While recent advancements in deep learning and large-scale datasets have driven progress, challenges such as data scarcity, cross-subject variability, and low semantic consistency persist. To address these issues, we introduce the concept of fMRI-to-Image Learning (fMRI2Image) and present the first systematic review in this field. This review highlights key challenges, categorizes methodologies such as fMRI signal encoding, feature mapping, and image generator. Finally, promising research directions are proposed to advance this emerging frontier, providing a reference for future studies.
\end{abstract}

\section{Introduction}
Functional magnetic resonance imaging (fMRI) is a powerful neuroimaging technique that measures brain activity indirectly by detecting changes in blood oxygen levels, which reflect neuronal activity. Recently, with the rise of deep learning models like Contrastive Language-Image Pre-training (CLIP) and Latent Diffusion Models (LDMs), along with the availability of large-scale fMRI datasets such as the Natural Scenes Dataset (NSD) \cite{NSD}, reconstructing visual perception from fMRI signals has become an exciting area of research. This approach not only enhances our understanding of how the brain encodes visual information but also opens new possibilities for applications like brain-computer interfaces (BCIs). By converting fMRI data into interpretable visual forms, we can explore the brain's internal representations and gain deeper insights into human perception and cognition.

Reconstructing visual images from fMRI signals presents several significant challenges, both in terms of data and modeling. Data-related challenges include the inherent complexity and variability of fMRI data, where individual differences in brain activity patterns can result in substantial variation across subjects. This variability complicates the creation of robust, generalized models. Moreover, fMRI datasets are often limited in diversity, particularly in terms of the number of subjects, which hinders the training of cross-subjects models. Another key issue is the misalignment of data across subjects, arising from the inherent variability in input dimensions due to differences in brain size. On the modeling side, mapping fMRI signals to the high-dimensional image space remains a challenging task. Effectively translating complex and noisy brain data into coherent visual representations requires advanced techniques that can capture subtle nuances in neural activity. Furthermore, while progress has been made in generating visual content from brain signals, issues with image quality and low-level semantic consistency persist. Current models still struggle to consistently generate images that accurately represent both fine details and the broader context of the visual stimuli. Additionally, traditional approaches often suffer from overfitting, particularly when trained on small datasets, as the models tend to memorize rather than generalize across different subjects and brain patterns. These challenges underscore the need for more sophisticated techniques to bridge the gap between fMRI and visual reconstructions.

To address the challenges of fMRI-to-image reconstruction, current methodologies are typically organized into three stages: fMRI signal encoding, feature mapping, and image reconstruction. Due to the varying dimensionalities and the presence of substantial noise in fMRI data across different subjects, many approaches convert fMRI data into one-dimensional representations of different lengths to reduce noise. Recently some methods, preserve spatial correlations by transforming the data into a two-dimensional standard brain map. One-dimensional data is typically processed using MLPs or transformers, while two-dimensional brain maps are often processed using CNNs or ViTs. Additionally, pretrained models on large neuroimaging datasets and techniques like Masked Autoencoders (MAE) further improve feature extraction, minimizing the need for paired fMRI-image datasets. The feature mapping stage aligns fMRI features with visual content by mapping features from different brain regions to corresponding embeddings. For instance, features from language-related brain areas are mapped to CLIP’s text embeddings. Finally, the image reconstruction stage generates visual content based on these aligned features. While earlier approaches struggled with low semantic consistency, recent advancements in diffusion models and Latent Diffusion Models (LDMs) like Stable Diffusion have significantly improved image quality and computational efficiency. Fine-tuning these models on specific fMRI datasets further addresses data scarcity and improves model generalization, highlighting the unique challenges of fMRI-to-image reconstruction compared to traditional text-to-image generation.

This review focuses on the latest advancements in reconstructing visual images from fMRI signals, a rapidly evolving interdisciplinary research area. It systematically organizes and classifies recent studies based on their methodologies and optimization objectives, while also highlighting the latest publicly available datasets that have facilitated progress in this field. 

\section{Datasets}
The fMRI-to-image (or-video) dataset facilitates the investigation of brain responses to static and dynamic visual stimuli. Data collection typically involves recruiting qualified participants from university communities. During experiments, participants perform continuous recognition tasks, viewing images of natural scenes or movie clips and indicating whether each stimulus has been previously encountered \cite{NSD}. The images are sourced from specific databases, while movie clips are selected from designated video databases \cite{VER}. High-field-strength fMRI scanners are employed to record participants' neural activity in response to visual stimuli.

Table \ref{dataset} provides an overview of the datasets, categorized by type, image/video source, number of subjects, sample size, and fMRI device specifications.

\begin{table*}[!t]
\tiny
\centering
\resizebox{\linewidth}{!}{
    \begin{tabular}{lcccccp{5.5cm}c} 
    \toprule
    \textbf{Dataset} & \textbf{Year}& \textbf{Type} & \textbf{Resource} & \textbf{Subject} & \textbf{Sample}  & \textbf{Device}    \\
    \midrule
    Vim-1&2008&Image & Corel Stock Photo Libraries/Berkeley Segmentation Dataset
 & 2 & 2*1750/120  & 4 T INOVA MR scanner/Quadrature transmit-receive surface coil   \\
    GOD&2015& Image & ImageNet & 5 & 5*1200/50  & 3.0-Tesla Siemens MAGNETOM Verio   \\ 
    EEG-VOA &2016& Image & ImageNet & 6 & 6*1600/400  & actiCAP-128Ch2/Brainvision DAQs   \\ 
    BOLD&2018& Image & ImageNet/SUN/COCO & 4 & 4*5254  & 3.0-Tesla Siemens MAGNETOM Verio   \\ 
    DIR&2017& Image & ImageNet & 3 & 3*1200/100  &3.0-Tesla Siemens MAGNETOM Verio   \\ 
    Faces&2018& Image & CelebA & 4 & 4*8000/20  & 3T Philips ACHIEVA scanner   \\ 
    largeEEG&2022& Image & THINGS & 10 & 82160 & 64-channel EASYCAP   \\ 
    OCD&2020& Image & ImageNet & 5 & 5*2250/500  & 3T Prismafit scanner   \\ 
    THINGS-data&2023& Image & THINGS & 3 & 3*8740  & Siemens 3T MAGNETOM Prisma/CTF 275 MEG system   \\ 
    NOD&2023& Image & ImageNet/COCO & 30 & 57120  & Siemens 3T MAGNETOM Prisma   \\ 
    NSD&2021& Image & COCO & 8 & 8*9000/1000 & 7T Siemens Magnetom 48 passively-shielded scanner/single-channel-transmit 32-channel-receive RF head coil   \\ 
    VER&2011& Video & Natural movies & 3 & 3*2400/180s & 4T Varian INOVA scanner   \\ 
    DNV&2016& Video & Natural movies & 3 & 3*972p & 3T MRI system/16-channel receive-only phase-array surface coil   \\ 
    STNS&2019& Video & Natural movies & 1 & 1*23h & Siemens 3T MAGNETOM Prisma   \\ 
    TGBH&2020& Video & The Grand Budapest Hotel & 25 & 25*2h & Siemens 3T MAGNETOM Prisma   \\ 
    NHA&2022& Video & Natural recording & 4 & 4*300p & 3T Siements TimTrio MR scanner/32-channel phase array head coil   \\ 
    NATVIEW-EEGfMRI &2023& Video & Checkerboard stimulus/Short film & 22 & 22*5958s & 3T Siements TimTrio MR scanner/MR-compatible system by Brain Products   \\ 
    BMD&2024& Video & Moments in Time & 10 & 10*1000p & 3T Trio Siemens scanner/32-channel head coil   \\ 
    m-fMRI&2024& Video & TSA2/UNBC-McMaster/Ganis \& Kievit/Polti & 101 & 101*6h & Siemens 3T MAGNETOM Prisma   \\ 
    NFED&2024& Video & DFEW/CAER & 5 & 5*1320p & Siemens 3T MAGNETOM Prisma   \\ 
    \bottomrule
    \end{tabular}
}
\caption{\textbf{Overview of Datasets.} In the Vim-1 dataset, for instance, the notation [2*1750/120] denotes that six participants viewed 1750 train images and 120 test images, yielding a total of 6*1870 samples. The absence of the \textbf{/} symbol indicates that no training-test split has been performed. The absence of the \textbf{*} indicates that the experimental conditions were different for each participant. For image datasets, the sample unit is the number of images. For video datasets, \textbf{s} denotes seconds, \textbf{h} denotes hours, and \textbf{p} denotes the number of videos.
}
\label{dataset}
\end{table*}

\subsection{fMRI to Image Datasets}
The earliest fMRI to image dataset can be traced back to the Vim-1 \cite{Vim-1} dataset proposed in 2008. The purpose of this research was to advance the understanding of how the brain represents visual information and to lay the foundation for potential future visual decoding technologies. In constructing this dataset, two healthy participants participated in the experiment. They viewed 1,750 natural images, including scenes and objects, to develop and test brain activity models based on fMRI data. To enhance the diversity of images, the Generic Object Decoding (GOD) data set \cite{GOD}, released in 2017, used natural images from 200 object categories within the ImageNet database. The Brain, Object, Landscape Dataset (BOLD) \cite{BOLD}, introduced in 2019, represents a milestone in visual research. This data set includes 5,254 images of real-world scenes from four participants, covering standard computer vision datasets from SUN, COCO, and ImageNet, thus significantly increasing sample diversity and scale. Currently, the Vim-1 dataset is often used alongside the GOD and BOLD datasets to evaluate models' generalization capabilities.

Previous fMRI-to-Image studies often used different image datasets than those in computer vision research, preventing the integration of neural data and computer vision models. The Deep Image Reconstruction (DIR) dataset \cite{DIR}, employing common computer vision image datasets, promotes cross-disciplinary research and model validation between neuroscience and computer vision. Distinguishing visually similar inputs, such as different instances within the same category or human faces, is challenging. In studies using the Face dataset \cite{Faces}, over 8,000 celebrity facial images were analyzed with deep learning models like Variational Autoencoders (VAE) combined with Generative Adversarial Networks (GAN) to capture complex facial features and subtle differences. The Object Category Decoding (OCD) dataset \cite{OCD} includes fMRI data from five healthy volunteers viewing five categories of natural images with 550 images per category. This dataset aids in studying brain decoding and processing of different object categories in natural scenes.

The Natural Scenes Dataset (NSD) \cite{NSD}, a large-scale 7T fMRI dataset, comprises high-resolution fMRI responses from 8 participants viewing 70,000+ unique, annotated natural scene images. Its scale, quality, and breadth make it a leading fMRI-to-image dataset. Subsequently, other high-quality, large-scale datasets emerged. The Natural Object Dataset (NOD) \cite{NOD}, comprising fMRI responses to 57,120 naturalistic images, is designed to minimize sampling variability. It enables the evaluation of inter-individual consistency and the generalizability of response patterns to diverse stimuli.
\subsection{fMRI to Video Datasets}
In addition to fMRI-to-image datasets focusing on static visual stimuli, recent studies have explored fMRI-to-video datasets utilizing dynamic stimuli such as short clips, movies, and natural scenes\cite{DNV,TGBH,NHA}. These datasets introduce a temporal dimension, offering new insights into how the brain processes complex, dynamic visual information.

\section{Methodology and Taxonomy}
\label{method}
\begin{figure*}[t]
    \centering
    \resizebox{\textwidth}{!}
    {
        \begin{forest}
            forked edges,
            for tree={
                grow=east,
                reversed=true,
                anchor=base west,
                parent anchor=east,
                child anchor=west,
                base=left,
                font=\small,
                rectangle,
                draw=hidden-draw,
                rounded corners,
                align=left,
                minimum width=2em,
                edge+={darkgray, line width=1pt},
                s sep=1pt,
                inner xsep=3pt,
                inner ysep=3pt,
                ver/.style={rotate=90, child anchor=north, parent anchor=south, anchor=center}
            },
            where level=1 {text width=4.9em, minimum height=2.0em, font=\fontsize{9}{5}\selectfont\bfseries}{},
            where level=2 {text width=6.9em, minimum height=2.2em, font=\fontsize{8}{8}\selectfont\bfseries}{},
            where level=3 {text width=5.7em, font=\fontsize{7}{8}\selectfont} {},
            [fMRI to Image Reconstruction, ver, color=mygrayp!\lcone, fill=mygrayp!15, text=black
                [\S \ref{task} \textbf{Method}, color=mygrayp!100, fill=mypurple!\lcone, text=black
                    [
                        \begin{tabular}{@{}l@{}l@{}}
                            \multirow{2}{*} {\S \ref{fMRI_encoding} ~} & \textbf{fMRI Signal} \\
                            & \textbf{Encoding} \\
                        \end{tabular}, color=mygrayp!100, fill=mypurple!\lctwo, text=black
                        [Architecture Design, color=mygrayp!100, fill=mypurple!\lcthree, text=black
                            [ {\textbf{1D:},\cite{DBLP:conf/icml/ScottiT0KCNSXNN24,DBLP:conf/nips/ScottiBGSNCDVYW23,wang2024mindbridge,DBLP:conf/icml/RadfordKHRGASAM21}\\
                            \cite{joo2024brainstreamsfmritoimagereconstructionmultimodal,Chen2024MindAC,wang2024mindbridge,DBLP:conf/icml/ScottiT0KCNSXNN24,quan2024psychometry}\\
                            \cite{takagi2023high,DBLP:conf/nips/ScottiBGSNCDVYW23,chen2023rethinking,Xia2024UMBRAEUM}}, prevent_leaf]
                            [{\textbf{2D:}\cite{huo2025neuropictor,gu2023decoding,lin2022mind,fang2020reconstructing,ferrante2022semantic}\\
                            \cite{meng2024semantics,DBLP:conf/mm/LuDZWH23}}, prevent_leaf] 
                        ]
                        [Encoder Pretrain, color=mygrayp!100, fill=mypurple!\lcthree, text=black
                            [{\cite{chen2023rethinking,qian2024lea,zeng2024controllable,huo2025neuropictor,ferrante2022semantic} \\  
                            \cite{ren2021reconstructing,ozcelik2022reconstruction,chen2023rethinking,10204983,gu2023decoding} \\
                            \cite{meng2024semantics,liu2024see,mai2023unibrain,chen2024cinematic}}, prevent_leaf]
                        ]
                    ]
                    [\begin{tabular}{@{}l@{}l@{}}
                        \multirow{2}{*}{\S \ref{Feature Mapping} ~}  & \textbf{Feature} \\
                        & \textbf{Mapping} \\
                    \end{tabular}, color=mygrayp!100, fill=mypurple!\lctwo, text=black
                        [Direct Alignment, color=mygrayp!100, fill=mypurple!\lcthree, text=black
                            [{\cite{Chen2024MindAC,Han2024MindFormerAT,ozcelik2023natural,huo2025neuropictor}}, prevent_leaf]
                        ]
                        [Indirect Alignment, color=mygrayp!100, fill=mypurple!\lcthree, text=black
                            [{\cite{gong2024litemind,lin2022mind,li2024neuraldiffuser,mai2023unibrain,gong2024neuroclips}}, prevent_leaf]
                        ]
                    ]
                    [\begin{tabular}{@{}l@{}l@{}}
                        \multirow{2}{*} {\S \ref{Image Reconstruction} ~} & \textbf{Image} \\
                        & \textbf{Generator} \\
                    \end{tabular}, color=mygrayp!100, fill=mypurple!\lctwo, text=black
                        [Partial Parameter \\ Fine-tuning, color=mygrayp!100, fill=mypurple!\lcthree, text=black
                            [{\cite{zeng2024controllable,10204983}}, prevent_leaf]
                        ]
                        [Incremental \\ Fine-tuning, color=mygrayp!100, fill=mypurple!\lcthree, text=black
                            [{\cite{guo2024mindldm,huo2025neuropictor,mai2023unibrain,balisacan2024neuro}}, prevent_leaf]
                        ]
                        [Result Inversion, color=mygrayp!100, fill=mypurple!\lcthree, text=black
                            [{\cite{DBLP:conf/mm/LuDZWH23,li2024neuraldiffuser}}, prevent_leaf]
                        ]
                    ]
                ]
                [\S \ref{task} \textbf{Task}, color=mygrayb!100, fill=myblue!\lcone, text=black
                    [\begin{tabular}{@{}l@{}l@{}}
                        \multirow{2}{*}{\S \ref{enhance reconstruction} ~}  & \textbf{Reconstruction}\\
                        & \textbf{Enhancement} \\
                    \end{tabular}, color=mygrayb!100, fill=myblue!\lctwo, text=black
                        [Low level, color=mygrayb!100, fill=myblue!\lcthree, text=black
                            [{\cite{joo2024brainstreamsfmritoimagereconstructionmultimodal,gu2023decoding,meng2023dual,gong2024neuroclips}\\
                            \cite{DBLP:conf/icml/ScottiT0KCNSXNN24,guo2024mindldm,gong2024mindtuner,li2024neuraldiffuser}\\
                            \cite{ozcelik2023natural,fang2020reconstructing,balisacan2024neuro}} , detect_leaf]    
                        ]
                        [Image quality, color=mygrayb!100, fill=myblue!\lcthree, text=black
                            [{\cite{zeng2024controllable,du2023decoding,shen2019end,ren2021reconstructing} }, detect_leaf]    
                        ] 
                    ]
                    [\begin{tabular}{@{}l@{}l@{}}
                        \multirow{2}{*}{\S \ref{model generalization} ~}  & \textbf{Model} \\
                        & \textbf{Generalization} \\
                    \end{tabular}, color=mygrayb!100, fill=myblue!\lctwo, text=black
                        [Cross subject, color=mygrayb!100, fill=myblue!\lcthree, text=black
                            [{\cite{wang2024mindbridge,DBLP:conf/icml/ScottiT0KCNSXNN24,Han2024MindFormerAT,jiang2024mindshot,gong2024mindtuner}\\
                            \cite{huo2025neuropictor,liu2024see,Ferrante2023ThroughTE,Xia2024UMBRAEUM}}, detect_leaf]    
                        ]
                        [Multi-task, color=mygrayb!100, fill=myblue!\lcthree, text=black
                            [{\cite{qian2024lea,gong2024litemind,Chen2024MindAC}\\
                            \cite{DBLP:conf/icml/ScottiT0KCNSXNN24,huo2025neuropictor,liu2024see}} , detect_leaf]    
                        ]
                        [Few-shot, color=mygrayb!100, fill=myblue!\lcthree, text=black
                            [{\cite{qian2024lea,gong2024litemind,Chen2024MindAC,wang2024mindbridge}\\
                            \cite{DBLP:conf/icml/ScottiT0KCNSXNN24,Han2024MindFormerAT,jiang2024mindshot,gong2024mindtuner}}, detect_leaf]    
                        ]
                        [fMRI to Video, color=mygrayb!100, fill=myblue!\lcthree, text=black
                            [{\cite{chen2024cinematic,gong2024neuroclips,lahner2024modeling,byrge2022video}}, detect_leaf]    
                        ]
                    ]
                ]
            ]
        \end{forest}}
    \caption{The main content flow and categorization of this survey.}
    \label{categorization_of_survey}
\end{figure*}
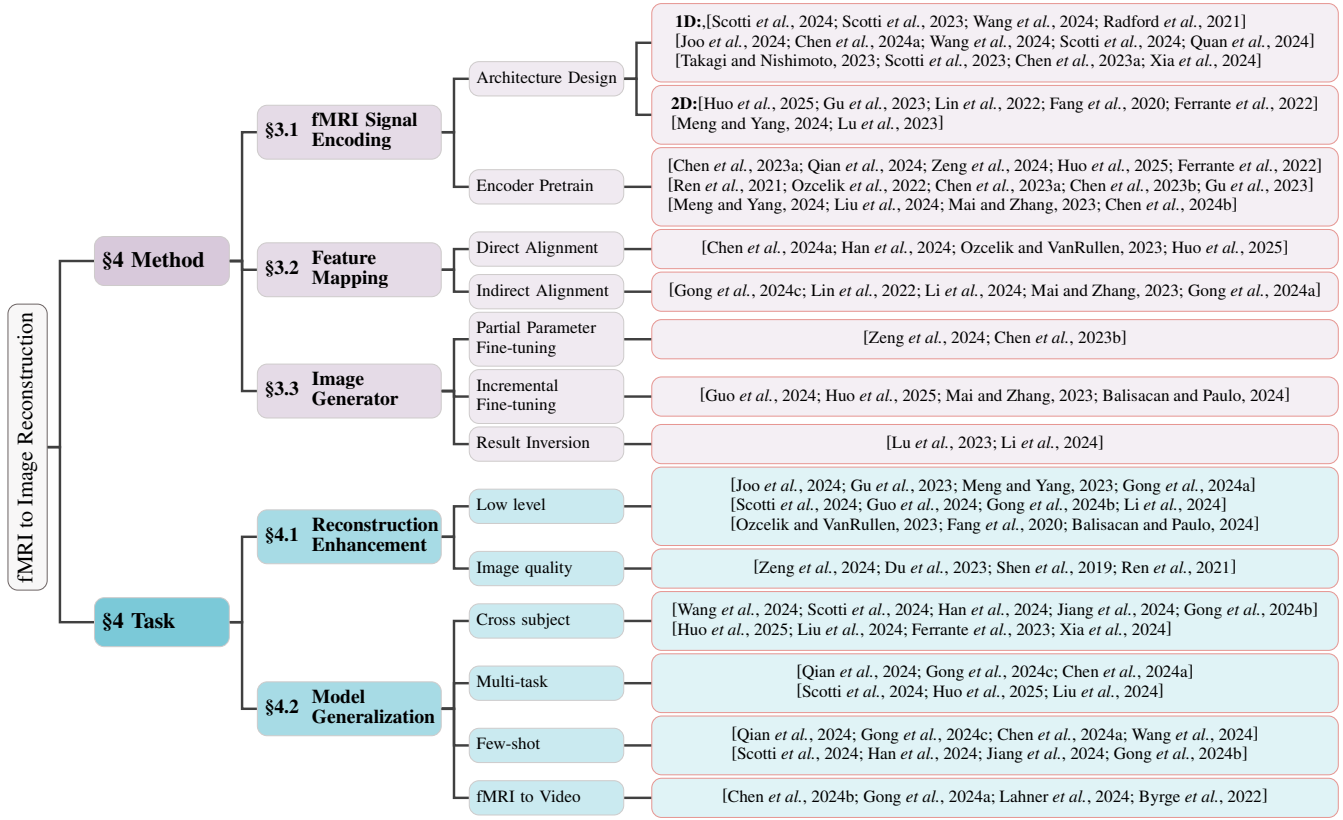

In this section, this paper synthesizes and examines the existing literature through the lens of model structure design, categorizing the fMRI image reconstruction method into approximately three modules as illustrated in Figure \ref{categorization_of_survey}:
(1). fMRI signal encoding: Encode the fMRI signal based on its data characteristics and abstract the features; (2). Feature alignment: Align the fMRI features with the existing modal features, such as CLIP \cite{DBLP:conf/icml/RadfordKHRGASAM21}; (3). Image reconstruction: Utilize the aligned fMRI features as conditional constraints for the generative model to guide the reconstruction of the original image.
Next, we will introduce these three modules individually.

\subsection{fMRI Signal Encoding}
\label{fMRI_encoding}
Encoding the fMRI signals is a fundamental and crucial step in the entire process, serving as the cornerstone for all subsequent modules.
It is essential to extract both high-level semantic information and low-level details, such as layout, color, and contour, from the original fMRI data in order to reconstruct the original image content.
The selection and configuration of the fMRI encoder are mainly determined by (1) architectural design and (2) the presence of pretraining.

\subsubsection{Architecture Design}

\textbf{1-D Architecture:} Many existing research approaches preprocess fMRI data into one-dimensional format through manual screening and other techniques to diminish data redundancy and noise interference~\cite{DBLP:conf/icml/ScottiT0KCNSXNN24,DBLP:conf/nips/ScottiBGSNCDVYW23,wang2024mindbridge,DBLP:conf/icml/RadfordKHRGASAM21}. Typically, the number of fMRI-image data pairs is limited, often in the tens of thousands~\cite{NSD}.
For such data, a straightforward approach would be to utilize simple networks like MLP for feature extraction~\cite{joo2024brainstreamsfmritoimagereconstructionmultimodal,meng2023dual,takagi2023high}. For instance, in Mindeye \cite{DBLP:conf/nips/ScottiBGSNCDVYW23}, MLP is employed for feature learning. In the follow-up study Mindeye2 \cite{DBLP:conf/icml/ScottiT0KCNSXNN24}, individual variances are addressed by training a distinct MLP for each participant.
One of its primary advantages lies in its simplicity and adaptability, making it highly suitable for the constrained data scale of fMRI. In scenarios involving multiple individuals, this approach can effectively address the requirements by employing a straightforward concept of one individual corresponding to one MLP~\cite{DBLP:conf/icml/ScottiT0KCNSXNN24}, without introducing excessive complexity.

\textbf{2-D Architecture:} Nevertheless, the drawbacks of basic 1-D networks like MLP are evident, particularly in their limited feature representation capabilities~\cite{dosovitskiy2021an,tolstikhin2021mlpmixer}, which hinder the comprehensive extraction of crucial information from complex data modalities such as fMRI. As a result, researchers have started exploring more sophisticated architectures, such as CNN networks, to enhance the representation of fMRI data and extract specific information from it.

Given that one-dimensional data lacks the spatial correlation present in brain signals, subsequent approaches like NeuroPictor \cite{huo2025neuropictor} have emerged to transform the processed data into a two-dimensional input format, preserving spatial information.
For this data format, researchers commonly employ ViT (vision transformer)~\cite{dosovitskiy2021an} as the foundational architecture to mimic the encoder architecture of CLIP \cite{DBLP:conf/icml/RadfordKHRGASAM21}, which enables the encoder to learn both global and local semantic information of fMRI, thereby enhancing the representation capabilities.
Nevertheless, a significant challenge with ViT is its demand for extensive training data, and the vulnerability of attention mechanisms to overfitting. This limitation is particularly pronounced in fMRI image reconstruction due to the scarcity of available data.

\subsubsection{Encoder Pretrain}

The constraint of limited data during encoder training has prompted researchers to adopt an alternative strategy, incorporating unpaired data for pre-training the fMRI encoder.
From the standpoint of the introduced data modality, the pre-training method can be categorized into two segments: (1) incorporating fMRI data and (2) integrating image data.

\textbf{Incorporate fMRI data.} In the first approach, researchers primarily embrace the concept of Masked Autoencoder (MAE)~\cite{MAE}. Through randomly masking fMRI data, they compel the encoder to grasp contextual information during training, thereby bolstering its representation capacity and reducing the data volume prerequisites for subsequent training on fMRI-image data pairs~\cite{qian2024lea,huo2025neuropictor,liu2024see}.
For instance, Chen et al. \cite{10204983} devise a masked brain modeling technique inspired by the MAE concept, leveraging the encoder for fMRI feature extraction.


\textbf{Incorporate image data.}
The second approach involves integrating extra image data~\cite{ren2021reconstructing,ozcelik2022reconstruction}. This methodology entails training the fMRI2Image encoder and Image2fMRI encoder independently, merging them to establish an Image-fMRI-Image process. This enables the utilization of image data for self-supervised training, thereby enhancing the encoder's representation capabilities. For instance, Gaziv et al. \cite{DBLP:conf/nips/BeliyGHSGI19} applied this concept by conducting supplementary training on the ImageNet image dataset to enhance the quality of reconstruction.
Both approaches involve integrating extra data to advance the representation capacity of fMRI data through self-supervised learning, finally reducing the need for fMRI-image data pairs in reconstruction tasks and enhancing the quality of reconstructions.
However, this strategy presents certain challenges. On one hand, it demands substantial computational resources due to the complexity of self-supervised learning frameworks. On the other hand, the observed improvements in image quality may partially result from the model, such as MAE, learning the specific styles inherent to the training dataset. This could limit the model’s ability to generalize to images with different styles, reducing its robustness across diverse datasets.

\subsection{Feature Mapping}
\label{Feature Mapping}
In the initial stages of fMRI2image reconstruction tasks, researchers typically decoded fMRI characteristics directly. More recent studies~\cite{guo2024mindldm,DBLP:conf/icml/ScottiT0KCNSXNN24} have started incorporating generative models like Diffusion~\cite{ho2020denoising} to enhance the reconstruction quality, necessitating the utilization of fMRI features as conditional constraints for the generative model.

Using the diffusion model as an illustration, this entails aligning fMRI features with their conditional constraint features, known as CLIP features. In terms of alignment approaches, we can broadly categorize them into two types: (1) direct alignment and (2) indirect alignment.

\textbf{Direct Alignment:} Direct alignment is highly intuitive as it directly aligns the fMRI features with the image signal itself or the CLIP features of the image description using MLP or linear layers. The primary objective of the linear layer and other components is to synchronize the token count and feature dimensions of the fMRI and Image features.
This alignment mainly employs Mean Squared Error (MSE) loss or contrastive loss \cite{DBLP:conf/icml/RadfordKHRGASAM21} to directly converge the absolute feature representations of the two entities in a specified distance space for fMRI semantic comprehension.
This method is nearly ubiquitous in recent studies focusing on diffusion-based reconstruction~\cite{jiang2024mindshot,huo2025neuropictor}. For instance, in MindBridge \cite{wang2024mindbridge}, two simulated features are produced via an encoder, which are aligned with the text and image characteristics of the CLIP, respectively. Brain-Streams \cite{joo2024brainstreamsfmritoimagereconstructionmultimodal} aligns fMRI features with standard features from other modalities at three distinct levels. Although this approach is direct and efficient, the low signal-to-noise ratio of fMRI and the substantial information mismatch between fMRI and image/text modalities frequently lead to a discrepancy in the features obtained through this direct method. The contrast loss setting, akin to that in CLIP, frequently need a larger batch size to manifest its advantages, leading to considerable resource overhead.

\textbf{Indirect Alignment:} Indirect alignment builds upon direct alignment by exploring the relative connection between fMRI and CLIP feature spaces, subsequently aligning this connection to enable fMRI features to converge towards the CLIP feature space. As an example, in CLIP-MUSED \cite{DBLP:conf/iclr/ZhouDWH24}, the first step involves computing the cosine distance between features in the fMRI space and CLIP space. Subsequently, it aligns features based on feature distances, thereby unveiling the intrinsic relationship within the CLIP space;
Lite-Mind \cite{gong2024litemind} operates by transforming all features into the frequency domain space, aligning features of distinct frequency domain components, and subsequently facilitating learning at various information levels. By focusing on relative alignments, the method enhances the model’s ability to capture subtle patterns and relationships that may not be evident through direct feature mapping. 

\subsection{Image Reconstruction}
\label{Image Reconstruction}
Initially, early fMRI2Image reconstruction efforts relied directly on fMRI features for decoding, leading to challenges in preserving semantic consistency and resulting in frequently nonsensical reconstruction outcomes. In recent years, the advancement of generative models has been notable, prompting researchers to leverage the capabilities of these models to enhance the completion of reconstruction tasks.
Prior to the introduction of Diffusion~\cite{ho2020denoising}, existing literature primarily focused on completing the reconstruction generation task using the GAN model~\cite{goodfellow2020generative}. The advent of diffusion has elevated image generation capabilities to a new height, resulting in a surge of fMRI image reconstruction work based on diffusion.
Due to the significant qualitative advancement offered by diffusion-based techniques compared to prior methods, this paper predominantly centers on diffusion-based approaches. Currently, the majority of research methodologies~\cite{wang2024mindbridge,gong2024mindtuner} directly employ fMRI features aligned with CLIP features to accomplish the reconstruction task by leveraging conditional constraints through diffusion.
Nevertheless, owing to the domain disparities between fMRI and CLIP data, constrained local perceptual capabilities inherent to CLIP, and the inconsistency with diffusion-based reconstruction outcomes, an increasing number of methodologies~\cite{huo2025neuropictor,10204983} opt to refine diffusion through fine-tuning.
Given the high cost associated with diffusion fine-tuning, contemporary fine-tuning processes typically adopt efficient strategies, broadly categorized into three types:

\textbf{Partial Parameter Fine-tuning:} Since fine-tuning all parameters is too costly and will destroy the original diffusion generation capability, researchers use partial fine-tuning, usually fine-tuning the cross-attention generated by conditional prompt intervention. For example, Chen et al. \cite{10204983} fine-tuned the prompt cross-attention layer in diffusion and added additional time step embedding to enhance the consistency of the reconstruction results.
This approach is straightforward and efficient; however, due to the constrained quantity of fine-tuning parameters, it may not entirely address the aforementioned issues.

\textbf{Incremental fine-tuning:} To enhance the fine-tuning effectiveness without compromising the inherent generation capability of diffusion, researchers shifted the fine-tuning focus from diffusion itself to an additional module~\cite{10204983}, thereby enabling diffusion fine-tuning while preserving its original generation capacity.
For instance, Zeng et al. \cite{zeng2024controllable} utilized the initial portion of the original diffusion decoder as a residual module for fine-tuning. They integrated the fine-tuning characteristics with the features of the original branch to accomplish the fine-tuning process.
The primary advantage of this approach lies in its capability to fine-tune the diffusion process without compromising the original generative capacity. This enables adaptation to the unique input characteristics of the fMRI modality and enhances the consistency of generated outcomes. Currently, this mode of thinking extends beyond fMRI image reconstruction tasks and finds widespread application in generation tasks, such as ControlNet \cite{DBLP:conf/iccv/ZhangRA23}.

\textbf{Result inversion:} The third approach to fine-tuning diffusion relies on the initialization state of diffusion. This method leverages the concept of inversion to directly minimize the disparity between the reconstructed output and the original image, enabling diffusion to identify more optimal initialization noise.
As an illustration, the MindDiffuser model \cite{DBLP:conf/mm/LuDZWH23} directly applies this concept to enhance the adaptability of diffusion to fMRI characteristics through fundamental CLIP feature alignment.
This method's advantage lies in its parameter-free nature, making it highly efficient and resource-light.
Nevertheless, as the core concept of the inversion approach is essentially data fine-tuning, akin to prompt learning, its applicability may be limited to specific types of fMRI features, such as the activation patterns of a particular individual or a specific semantic category.

\section{Optimization Objective}
\label{task}
The current optimization objective of fMRI2Image methods can be broadly categorized into reconstruction enhancement and model generalization enhancement. Reconstruction enhancement primarily focuses on two aspects: (1) improving the quality of generated images by optimizing the image generation module, and (2) enhancing the reconstruction of image details through optimization of the representation or alignment methods. Model generalization enhancement addresses issues related to poor generalization caused by factors such as limited data and large individual differences. Key tasks in this area include cross-subject model, multi-task learning, and few-shot learning.

\subsection{Enhance reconstruction}
\label{enhance reconstruction}
\textbf{Low level:} To reconstruct complex natural images from fMRI data, hierarchical cues—such as precise spatial layouts and semantic details—must be extracted from corresponding brain regions. Various expressive pre-trained models are leveraged to learn semantic and fine-grained image features for better alignment with fMRI signals. For instance, Cortex2Image~\cite{gu2023decoding} employs SwAV to extract ground-truth semantic vectors and uses a variational approach to capture fine-grained image details.
To further enhance semantic feature representation, some methods introduce additional information, such as depth cues. Takagi et al.~\cite{takagi2023high} integrate depth information by aligning predicted depth from brain activity with the latent representation of the DPT model from Hugging Face, subsequently feeding it into a Stable Diffusion (SD) model along with semantic features for image reconstruction. In addition to image features, the absence of textual descriptions in datasets like GOD (derived from ImageNet) poses a challenge. To address this, Meng et al.\cite{meng2023dual} propose the Dual-Guided Brain Diffusion Model (DBDM), which uses BLIP to generate captions for training images, followed by semantic feature extraction via a CLIP text encoder. Expanding on this, BrainStreams\cite{joo2024brainstreamsfmritoimagereconstructionmultimodal} incorporates multi-modal guidance at three levels. Leveraging the two-streams hypothesis, it adopts a brain region-specific approach to separately extract semantic and perceptual information from fMRI data. A large language model refines predicted captions, aligning them with BERT’s latent vectors of annotations, while mid- and low-level guidance is provided through CLIP image embeddings and SD latent vectors.

\textbf{Image quality:}  Enhancing image quality is one of the most common tasks in fMRI-to-image reconstruction. The goal is to improve the overall quality of the generated images, such as their similarity to the original images, as well as the coherence of content and color.
Common approaches in this field include:
IC-GAN \cite{ozcelik2022reconstruction} extracts instance features, noise vectors, and dense vectors from training images and uses ridge regression models to predict these latent variables from fMRI patterns. By conditioning image generation on these predicted variables, it improves semantic attributes and preserves the coherence of content. MinD-Vis \cite{10204983}employs masked brain modeling to learn effective self-supervised representations of fMRI data. With a double-conditioned latent diffusion model, it generates plausible images with semantically matching details, outperforming previous methods in semantic mapping and generation quality. VQ-fMRI \cite{chen2023rethinking} formulates visual reconstruction as experience-based context completion, guided by visual cues from brain activities. It learns discrete visual representations and constituent contexts in a self-supervised manner, utilizing a token-to-token inpainting network to complete visual content, significantly enhancing the quality of reconstructed images, particularly in color and texture.

\subsection{Model generalization}
\label{model generalization}
\textbf{Cross-subject:} In the realm of fMRI2Image reconstruction, existing methods focus on training models on a per-subject basis. Consequently, models trained on fMRI data from a specific individual are typically restricted to that same individual. The challenges in cross-subject optimization mainly lie in the inherent differences of human brain. Different brain size may cause significant differences in the shape of fMRI data collected. Additionally, the neural responses vary from subject to subject due to their individual experiences and biases, making it hard to achieve generalized latent representation of brain signals from different subject. MindEye2\cite{DBLP:conf/icml/ScottiT0KCNSXNN24} resolves the problem of shape difference by leveraging an initial alignment step to handle input from different subjects, where their fMRI voxels are projected into a shared latent space through a separate linear layer. MindFormer\cite{Han2024MindFormerAT} incorporates a unique subject token through a subsequent transformer encoder to the output of linear layer in the shared latent space, enhancing the interpretation accuracy of diverse neural response patterns. MindBridge\cite{wang2024mindbridge} further integrates AutoEncoder and cyclic mechanism to simulate two subjects viewing the same stimuli images, adding a loss minimizing their distance to the training pipeline to learn subject-invariant semantic embeddings.

\textbf{Multi-task:} In the field of fMRI2Image reconstruction, the task of multi-task learning involves training a model to perform multiple related tasks simultaneously. This approach aims to leverage shared information among different tasks to improve the overall performance and generalization ability of the model. By jointly optimizing for multiple tasks, the model can learn more comprehensive and meaningful representations of the fMRI data, leading to enhanced performance in various related subtasks. Challenges in multi-task fMRI-to-image conversion include effectively balancing the learning of different tasks to avoid overfitting or underfitting on any single task. Additionally, handling the diverse nature of the tasks, such as image reconstruction, retrieval, and classification, requires careful design of the model architecture and training strategy to ensure that the model can capture the specific characteristics and requirements of each task. Liu et al. \cite{liu2024see} proposes a neural decoding model that combines a high-level perception decoding pipeline and a pixel-wise reconstruction pipeline. It uses contrastive learning to align fMRI data with visual and textual modalities, enabling tasks such as fMRI-to-image retrieval, fMRI-to-text retrieval, zero-shot classification, and fMRI-to-image generation. The model is trained on data from multiple subjects to learn shared response patterns and capture individual-level deviations, enhancing its generalization ability across different tasks. Lite-Mind \cite{gong2024litemind} focuses on fMRI-to-image retrieval. It designs a DFT Backbone with Spectrum Compression and Frequency Projector modules to learn informative and robust voxel embeddings. By efficiently aligning fMRI voxels to the fine-grained information of CLIP, Lite-Mind achieves high retrieval accuracy with significantly fewer parameters compared to previous methods. It can be applied to different downstream tasks such as zero-shot classification, demonstrating its versatility in handling multiple related tasks. NeuroPictor \cite{huo2025neuropictor} divides the fMRI-to-image process into three steps. It first learns a universal latent fMRI space through multi-individual pre-training to capture signal information and individual differences. Then, it extracts high-level semantic and low-level structure features from the latent fMRI to guide the generation process of the diffusion model. This method enables precise control over image creation, achieving high-quality reconstructions and performing well in both fMRI decoding and encoding tasks, thus handling multiple aspects of the fMRI-to-image conversion process. LEA(Joint fMRI Decoding and Encoding with Latent Embedding Alignment) \cite{qian2024lea} constructs latent spaces for fMRI signals and images and aligns them to enable bidirectional transformation. It uses an encoder-decoder architecture for each modality and an alignment module to connect the latent spaces. This allows the model to perform both neural decoding (recovering visual stimuli from fMRI signals) and neural encoding (predicting brain activity from images) tasks within a unified framework. LEA addresses the challenges of fMRI data, such as redundancy, instability, and insufficiency, and produces high-fidelity semantic-consistent results in multiple tasks. MindEye2 \cite{DBLP:conf/icml/ScottiT0KCNSXNN24} pretrains a model across multiple subjects and then fine-tunes it on limited data from a new subject. It maps fMRI activity to a shared-subject latent space using ridge regression and then to the CLIP image space. The model reconstructs images with the help of a fine-tuned Stable Diffusion XL unCLIP model and also predicts image captions. It achieves state-of-the-art performance in fMRI-to-image reconstruction and retrieval metrics and demonstrates the ability to handle tasks such as image captioning and brain correlation analysis in addition to image reconstruction.

\textbf{Few-shot:} In fMRI-to-image reconstruction, acquiring a large amount of fMRI-image paired data is extremely difficult and time-consuming. The limited training data makes it challenging for models to learn effective mappings from brain activity to visual stimuli, often leading to overfitting or poor generalization. To address the data scarcity problem, researchers employ few-shot learning strategies.
MindShot\cite{jiang2024mindshot} proposes a Fourier-based cross-subject supervision framework. It first uses contrastive learning to pretrain on multiple subjects to obtain prior knowledge. Then, for new subjects, it applies an HRF adapter to correct individual differences. By using Fourier transform, it extracts high-level and low-level features from other subjects' signals for cross-subject supervision. This approach enables the model to achieve effective few-shot brain decoding and outperforms per-subject-per-model paradigms, especially in scenarios with very limited data. MindEye2\cite{DBLP:conf/icml/ScottiT0KCNSXNN24} is pretrained on data from 7 subjects and then fine-tuned on a new subject with minimal data (as little as 1 hour of scanning). It uses a novel functional alignment procedure with subject-specific ridge regression to map fMRI activity to a shared-subject latent space. By leveraging this shared-space approach and fine-tuning on limited data, it can achieve high-quality reconstructions and competitive decoding performance even with a small amount of training data from the new subject.
Lite-Mind \cite{gong2024litemind} uses Discrete Fourier Transform (DFT) to process fMRI signals. It designs a DFT backbone with Spectrum Compression and Frequency Projector modules to learn robust voxel embeddings. This method is highly efficient and can achieve good results with a relatively small amount of data. For example, it achieves high fMRI-to-image retrieval accuracy on the NSD dataset with significantly fewer parameters compared to other models, demonstrating its effectiveness in handling limited data scenarios.These innovative approaches highlight the potential of leveraging frequency-domain transformations and cross-subject learning to overcome data limitations, paving the way for more generalizable and efficient brain decoding models.

\textbf{fMRI to Video:} fMRI-to-video reconstruction is a complex task that requires capturing the temporal dynamics and continuity of visual experiences. The objective is to generate videos with high visual quality, semantic consistency, and smooth frame transitions. Currently, common approaches in this field include:
Progressive Learning Approaches—MinD-Video adopts this strategy, leveraging masked brain modeling, multi-modal contrastive learning, and co-training with an augmented Stable Diffusion model to produce high-quality videos with precise semantics and dynamics, outperforming previous techniques \cite{chen2024cinematic}. Unified Frameworks with Multi-Modal Information—NeuroCLIPs focuses on high-fidelity video reconstruction using a unified framework that combines visual and textual information. Through a two-stage training process, it enhances visual quality and semantic consistency, achieving notable improvements over earlier methods \cite{gong2024neuroclips}.

\section{Conclusion and Future Trends}
In summary, this paper provides a comprehensive review of the fMRI-to-image reconstruction process. The existing literature is then categorized into three main areas: fMRI signal encoder design, feature mapping, and image reconstruction. Additionally, we highlight six key questions that are central to the field: low-level image reconstruction, image quality, cross-subject variability, multi-task learning, few-shot learning, and fMRI2video. For each of these areas, we present representative methodologies and discuss their key technical contributions. Despite the progress made, several unresolved challenges remain, indicating the need for continued research and innovation in this domain.
\paragraph{Generalization to New Subjects:}A significant challenge in fMRI-to-image reconstruction is the ability to generalize across subjects. Current models often rely on subject-specific data, which limits their applicability to new individuals or leads to the forgetting of information from previous subjects. Future research should focus on developing more robust methods that can generalize well across different subjects, accounting for the inherent variability in brain activity patterns. Techniques such as transfer learning and domain adaptation could help address this challenge, enabling the creation of models that are both subject-independent and scalable to to larger, more diverse populations.
\paragraph{Interpretability and Explainability:}As machine learning models become increasingly complex, the need for interpretability and explainability in fMRI2Image reconstruction is growing. One promising direction is to explore attention mechanisms and other explainable AI techniques to better understand the relationship between specific brain regions and the generated content. By identifying which brain areas are activated during specific tasks or stimuli, researchers could gain deeper insights into the neural processes underlying perception and cognition. Such advancements could also improve trust in AI-driven neuroimaging applications, making them more transparent and clinically applicable.
\paragraph{Direct Video Generation from fMRI:}Another exciting frontier is the direct generation of video or dynamic content from fMRI data. While current methods focus mainly on generating still images, the temporal resolution of fMRI, although lower than that of EEG, is still sufficient to capture key patterns of brain activity over time. By leveraging this temporal dimension, future models could potentially reconstruct dynamic visual content, including videos or even real-time brain activity visualizations. This would open up new possibilities for studying dynamic brain processes, as well as applications in virtual reality, neuroscience research, and brain-computer interfaces.
\bibliographystyle{named}
\bibliography{ijcai25}

\begin{thebibliography}{}

\bibitem[\protect\citeauthoryear{Allen and St-Yves}{2021}]{NSD}
E.J. Allen and G.~St-Yves.
\newblock A massive 7t fmri dataset to bridge cognitive neuroscience and artificial intelligence.
\newblock {\em Nature Neuroscience}, 2021.

\bibitem[\protect\citeauthoryear{Balisacan and Paulo}{2024}]{balisacan2024neuro}
G.M. Balisacan and A.T.A. Paulo.
\newblock Neuro-vis: Guided complex image reconstruction from brain signals using multiple semantic and perceptual controls.
\newblock In {\em ICML}, pages 1--8, 2024.

\bibitem[\protect\citeauthoryear{Beliy \bgroup \em et al.\egroup }{2019}]{DBLP:conf/nips/BeliyGHSGI19}
R.~Beliy, G.~Gaziv, A.~Hoogi, F.~Strappini, T.~Golan, and M.~Irani.
\newblock From voxels to pixels and back: Self-supervision in natural-image reconstruction from fmri.
\newblock In {\em NIPS}, pages 6514--6524, 2019.

\bibitem[\protect\citeauthoryear{Byrge \bgroup \em et al.\egroup }{2022}]{byrge2022video}
L.~Byrge, D.~Kliemann, Y.~He, H.~Cheng, J.M. Tyszka, R.~Adolphs, and D.P. Kennedy.
\newblock Video-evoked fmri bold responses are highly consistent across different data acquisition sites.
\newblock {\em Human brain mapping}, 2022.

\bibitem[\protect\citeauthoryear{Castello and Chauhan}{2020}]{TGBH}
M.~Castello and V.~Chauhan.
\newblock An fmri dataset in response to “the grand budapest hotel”, a socially-rich, naturalistic movie.
\newblock {\em Scientific Data}, 7, 2020.

\bibitem[\protect\citeauthoryear{Chang and Pyles}{2018}]{BOLD}
N.~Chang and J.A. Pyles.
\newblock Bold5000, a public fmri dataset while viewing 5000 visual images.
\newblock {\em Scientific Data}, 6, 2018.

\bibitem[\protect\citeauthoryear{Chen \bgroup \em et al.\egroup }{2023a}]{chen2023rethinking}
J.~Chen, Y.~Qi, and G.~Pan.
\newblock Rethinking visual reconstruction: experience-based content completion guided by visual cues.
\newblock In {\em ICML}, 2023.

\bibitem[\protect\citeauthoryear{Chen \bgroup \em et al.\egroup }{2023b}]{10204983}
Z.~Chen, J.~Qing, T.~Xiang, W.L. Yue, and J.H. Zhou.
\newblock Seeing beyond the brain: Conditional diffusion model with sparse masked modeling for vision decoding.
\newblock In {\em CVPR}, pages 22710--22720, 2023.

\bibitem[\protect\citeauthoryear{Chen \bgroup \em et al.\egroup }{2024a}]{Chen2024MindAC}
J.~Chen, Y.~Qi, Y.~Wang, and G.~Pan.
\newblock Mind artist: Creating artistic snapshots with human thought.
\newblock {\em CVPR}, pages 27197--27207, 2024.

\bibitem[\protect\citeauthoryear{Chen \bgroup \em et al.\egroup }{2024b}]{chen2024cinematic}
Z.~Chen, J.~Qing, and J.H. Zhou.
\newblock Cinematic mindscapes: High-quality video reconstruction from brain activity.
\newblock {\em NIPS}, 36, 2024.

\bibitem[\protect\citeauthoryear{Dosovitskiy \bgroup \em et al.\egroup }{2021}]{dosovitskiy2021an}
A.~Dosovitskiy, L.~Beyer, A.~Kolesnikov, D.~Weissenborn, X.~Zhai, T.~Unterthiner, M.~Dehghani, M.~Minderer, G.~Heigold, S.~Gelly, J.~Uszkoreit, and N.~Houlsby.
\newblock An image is worth 16x16 words: Transformers for image recognition at scale.
\newblock In {\em ICLR}, 2021.

\bibitem[\protect\citeauthoryear{Du \bgroup \em et al.\egroup }{2023}]{du2023decoding}
C.~Du, K.~Fu, J.~Li, and H.~He.
\newblock Decoding visual neural representations by multimodal learning of brain-visual-linguistic features.
\newblock {\em TPAMI}, 2023.

\bibitem[\protect\citeauthoryear{Fang \bgroup \em et al.\egroup }{2020}]{fang2020reconstructing}
T.~Fang, Y.~Qi, and G.~Pan.
\newblock Reconstructing perceptive images from brain activity by shape-semantic gan.
\newblock {\em NIPS}, 33:13038--13048, 2020.

\bibitem[\protect\citeauthoryear{Ferrante \bgroup \em et al.\egroup }{2022}]{ferrante2022semantic}
M.~Ferrante, T.~Boccato, and N.~Toschi.
\newblock Semantic brain decoding: from fmri to conceptually similar image reconstruction of visual stimuli.
\newblock {\em arXiv preprint arXiv:2212.06726}, 2022.

\bibitem[\protect\citeauthoryear{Ferrante \bgroup \em et al.\egroup }{2023}]{Ferrante2023ThroughTE}
M.~Ferrante, T.~Boccato, and N.~Toschi.
\newblock Through their eyes: Multi-subject brain decoding with simple alignment techniques.
\newblock {\em Imaging Neuroscience}, 2:1--21, 2023.

\bibitem[\protect\citeauthoryear{Gong and Zhou}{2023}]{NOD}
Z.~Gong and M.~Zhou.
\newblock A large-scale fmri dataset for the visual processing of naturalistic scenes.
\newblock {\em Scientific Data}, 10, 2023.

\bibitem[\protect\citeauthoryear{Gong \bgroup \em et al.\egroup }{2024a}]{gong2024neuroclips}
Z.~Gong, G.~Bao, Q.~Zhang, Z.~Wan, D.~Miao, S.~Wang, L.~Zhu, C.~Wang, R.~Xu, and L.~Hu.
\newblock Neuroclips: Towards high-fidelity and smooth fmri-to-video reconstruction.
\newblock {\em arXiv:2410.19452}, 2024.

\bibitem[\protect\citeauthoryear{Gong \bgroup \em et al.\egroup }{2024b}]{gong2024mindtuner}
Z.~Gong, Q.~Zhang, G.~Bao, L.~Zhu, K.~Liu, L.~Hu, and D.~Miao.
\newblock Mindtuner: Cross-subject visual decoding with visual fingerprint and semantic correction.
\newblock {\em arXiv preprint arXiv:2404.12630}, 2024.

\bibitem[\protect\citeauthoryear{Gong \bgroup \em et al.\egroup }{2024c}]{gong2024litemind}
Z.~Gong, Q.~Zhang, G.~Bao, L.~Zhu, Y.~Zhang, K.~Liu, L.~Hu, and D.~Miao.
\newblock Lite-mind: Towards efficient and robust brain representation learning.
\newblock In {\em MM}, 2024.

\bibitem[\protect\citeauthoryear{Goodfellow \bgroup \em et al.\egroup }{2020}]{goodfellow2020generative}
I.~Goodfellow, J.~Pouget-Abadie, M.~Mirza, B.~Xu, D.~Warde-Farley, S.~Ozair, A.~Courville, and Y.~Bengio.
\newblock Generative adversarial networks.
\newblock {\em Communications of the ACM}, 63(11):139--144, 2020.

\bibitem[\protect\citeauthoryear{Gu \bgroup \em et al.\egroup }{2023}]{gu2023decoding}
Z.~Gu, K.~Jamison, A.~Kuceyeski, and M.~Sabuncu.
\newblock Decoding natural image stimuli from fmri data with a surface-based convolutional network.
\newblock {\em MIDL}, 2023.

\bibitem[\protect\citeauthoryear{Guo \bgroup \em et al.\egroup }{2024}]{guo2024mindldm}
J.~Guo, C.~Yi, F.~Li, P.~Xu, and Y.~Tian.
\newblock Mindldm: Reconstruct visual stimuli from fmri using latent diffusion model.
\newblock In {\em CIVEMSA}. IEEE, 2024.

\bibitem[\protect\citeauthoryear{Han \bgroup \em et al.\egroup }{2024}]{Han2024MindFormerAT}
I.~Han, J.~Lee, and J.C. Ye.
\newblock Mindformer: A transformer architecture for multi-subject brain decoding via fmri.
\newblock {\em ArXiv}, abs/2405.17720, 2024.

\bibitem[\protect\citeauthoryear{He \bgroup \em et al.\egroup }{2022}]{MAE}
K.~He, X.~Chen, S.~Xie, Y.~Li, P.~Dollár, and R.~Girshick.
\newblock Masked autoencoders are scalable vision learners.
\newblock In {\em CVPR}, pages 15979--15988, 2022.

\bibitem[\protect\citeauthoryear{Ho \bgroup \em et al.\egroup }{2020}]{ho2020denoising}
J.~Ho, A.~Jain, and P.~Abbeel.
\newblock Denoising diffusion probabilistic models.
\newblock In {\em NIPS}, 2020.

\bibitem[\protect\citeauthoryear{Horikawa and Kamitani}{2015}]{GOD}
T.~Horikawa and Y.~Kamitani.
\newblock Generic decoding of seen and imagined objects using hierarchical visual features.
\newblock {\em Nature Communications}, 2015.

\bibitem[\protect\citeauthoryear{Huang and Yan}{2020}]{OCD}
W.~Huang and H.~Yan.
\newblock Long short‐term memory‐based neural decoding of object categories evoked by natural images.
\newblock {\em Human Brain Mapping}, 41:4442 -- 4453, 2020.

\bibitem[\protect\citeauthoryear{Huo \bgroup \em et al.\egroup }{2025}]{huo2025neuropictor}
J.~Huo, Y.~Wang, Y.~Wang, X.~Qian, C.~Li, Y.~Fu, and J.~Feng.
\newblock Neuropictor: Refining fmri-to-image reconstruction via multi-individual pretraining and multi-level modulation.
\newblock In {\em ECCV}, pages 56--73. Springer, 2025.

\bibitem[\protect\citeauthoryear{Jiang \bgroup \em et al.\egroup }{2024}]{jiang2024mindshot}
S.~Jiang, Z.~Meng, D.~Liu, H.~Li, F.~Su, and Z.~Zhao.
\newblock Mindshot: Brain decoding framework using only one image.
\newblock {\em arXiv preprint arXiv:2405.15278}, 2024.

\bibitem[\protect\citeauthoryear{Joo \bgroup \em et al.\egroup }{2024}]{joo2024brainstreamsfmritoimagereconstructionmultimodal}
J.~Joo, T.~Jeong, and S.~Hwang.
\newblock Brain-streams: fmri-to-image reconstruction with multi-modal guidance.
\newblock In {\em arxiv}, 2024.

\bibitem[\protect\citeauthoryear{Kay and Naselaris}{2008}]{Vim-1}
K.N. Kay and T.~Naselaris.
\newblock Identifying natural images from human brain activity.
\newblock {\em Nature}, 452:352--355, 2008.

\bibitem[\protect\citeauthoryear{Lahner \bgroup \em et al.\egroup }{2024}]{lahner2024modeling}
B.~Lahner, K.~Dwivedi, P.~Iamshchinina, M.~Graumann, A.~Lascelles, G.~Roig, A.T. Gifford, B.~Pan, S.~Jin, and N.A.~Ratan Murty.
\newblock Modeling short visual events through the bold moments video fmri dataset and metadata.
\newblock {\em Nature communications}, 15(1):6241, 2024.

\bibitem[\protect\citeauthoryear{Li \bgroup \em et al.\egroup }{2024}]{li2024neuraldiffuser}
H.~Li, H.~Wu, and B.~Chen.
\newblock Neuraldiffuser: Controllable fmri reconstruction with primary visual feature guided diffusion.
\newblock {\em arXiv:2402.13809}, 2024.

\bibitem[\protect\citeauthoryear{Lin \bgroup \em et al.\egroup }{2022}]{lin2022mind}
S.~Lin, T.~Sprague, and A.K. Singh.
\newblock Mind reader: Reconstructing complex images from brain activities.
\newblock {\em NIPS}, 35:29624--29636, 2022.

\bibitem[\protect\citeauthoryear{Liu \bgroup \em et al.\egroup }{2024}]{liu2024see}
Y.~Liu, Y.~Ma, G.~Zhu, H.~Jing, and N.~Zheng.
\newblock See through their minds: Learning transferable neural representation from cross-subject fmri.
\newblock {\em arXiv preprint arXiv:2403.06361}, 2024.

\bibitem[\protect\citeauthoryear{Lu \bgroup \em et al.\egroup }{2023}]{DBLP:conf/mm/LuDZWH23}
Y.~Lu, C.~Du, Q.~Zhou, D.~Wang, and H.~He.
\newblock Minddiffuser: Controlled image reconstruction from human brain activity with semantic and structural diffusion.
\newblock In {\em MM}, pages 5899--5908. {ACM}, 2023.

\bibitem[\protect\citeauthoryear{Mai and Zhang}{2023}]{mai2023unibrain}
W.~Mai and Z.~Zhang.
\newblock Unibrain: Unify image reconstruction and captioning all in one diffusion model from human brain activity.
\newblock {\em arXiv preprint arXiv:2308.07428}, 2023.

\bibitem[\protect\citeauthoryear{Meng and Yang}{2023}]{meng2023dual}
L.~Meng and C.~Yang.
\newblock Dual-guided brain diffusion model: Natural image reconstruction from human visual stimulus fmri.
\newblock {\em Bioengineering}, 2023.

\bibitem[\protect\citeauthoryear{Meng and Yang}{2024}]{meng2024semantics}
L.~Meng and C.~Yang.
\newblock Semantics-guided hierarchical feature encoding generative adversarial network for visual image reconstruction from brain activity.
\newblock {\em IEEE TNSRE}, 2024.

\bibitem[\protect\citeauthoryear{Nishimoto and Vu}{2011}]{VER}
S.~Nishimoto and A.T. Vu.
\newblock Reconstructing visual experiences from brain activity evoked by natural movies.
\newblock {\em Current Biology}, 21:1641--1646, 2011.

\bibitem[\protect\citeauthoryear{Ozcelik and VanRullen}{2023}]{ozcelik2023natural}
F.~Ozcelik and R.~VanRullen.
\newblock Natural scene reconstruction from fmri signals using generative latent diffusion.
\newblock {\em Scientific Reports}, 2023.

\bibitem[\protect\citeauthoryear{Ozcelik \bgroup \em et al.\egroup }{2022}]{ozcelik2022reconstruction}
F.~Ozcelik, B.~Choksi, M.~Mozafari, L.~Reddy, and R.~VanRullen.
\newblock Reconstruction of perceived images from fmri patterns and semantic brain exploration using instance-conditioned gans.
\newblock In {\em IJCNN}. IEEE, 2022.

\bibitem[\protect\citeauthoryear{Qian \bgroup \em et al.\egroup }{2024}]{qian2024lea}
X.~Qian, Y.~Wang, X.~Sun, Y.~Fu, X.~Xue, and J.~Feng.
\newblock {LEA}: Learning latent embedding alignment model for f{MRI} decoding and encoding.
\newblock {\em TMLR}, 2024.

\bibitem[\protect\citeauthoryear{Quan \bgroup \em et al.\egroup }{2024}]{quan2024psychometry}
R.~Quan, W.~Wang, Z.~Tian, F.~Ma, and Y.~Yang.
\newblock Psychometry: An omnifit model for image reconstruction from human brain activity.
\newblock In {\em CVPR}, 2024.

\bibitem[\protect\citeauthoryear{Radford \bgroup \em et al.\egroup }{2021}]{DBLP:conf/icml/RadfordKHRGASAM21}
A.~Radford, J.W. Kim, C.~Hallacy, A.~Ramesh, G.~Goh, S.~Agarwal, G.~Sastry, A.~Askell, P.~Mishkin, J.~Clark, G.~Krueger, and I.~Sutskever.
\newblock Learning transferable visual models from natural language supervision.
\newblock In {\em ICML}, volume 139 of {\em Proceedings of Machine Learning Research}, pages 8748--8763. {PMLR}, 2021.

\bibitem[\protect\citeauthoryear{Ren \bgroup \em et al.\egroup }{2021}]{ren2021reconstructing}
Z.~Ren, J.~Li, X.~Xue, X.~Li, F.~Yang, Z.~Jiao, and X.~Gao.
\newblock Reconstructing seen image from brain activity by visually-guided cognitive representation and adversarial learning.
\newblock {\em NeuroImage}, 228:117602, 2021.

\bibitem[\protect\citeauthoryear{Scotti \bgroup \em et al.\egroup }{2023}]{DBLP:conf/nips/ScottiBGSNCDVYW23}
P.S. Scotti, A.~Banerjee, J.~Goode, S.~Shabalin, A.~Nguyen, E.~Cohen, A.J. Dempster, N.~Verlinde, E.~Yundler, D.~Weisberg, K.A. Norman, and T.M. Abraham.
\newblock Reconstructing the mind's eye: fmri-to-image with contrastive learning and diffusion priors.
\newblock In {\em NIPS}, 2023.

\bibitem[\protect\citeauthoryear{Scotti \bgroup \em et al.\egroup }{2024}]{DBLP:conf/icml/ScottiT0KCNSXNN24}
P.S. Scotti, M.~Tripathy, C.~Torrico, R.~Kneeland, T.~Chen, A.~Narang, C.~Santhirasegaran, J.~Xu, T.~Naselaris, K.A. Norman, and T.M. Abraham.
\newblock Mindeye2: Shared-subject models enable fmri-to-image with 1 hour of data.
\newblock In {\em ICML}. OpenReview.net, 2024.

\bibitem[\protect\citeauthoryear{Shen and Horikawa}{2017}]{DIR}
G.~Shen and T.~Horikawa.
\newblock Deep image reconstruction from human brain activity.
\newblock {\em PLoS Computational Biology}, 15, 2017.

\bibitem[\protect\citeauthoryear{Shen \bgroup \em et al.\egroup }{2019}]{shen2019end}
G.~Shen, K.~Dwivedi, K.~Majima, T.~Horikawa, and Y.~Kamitani.
\newblock End-to-end deep image reconstruction from human brain activity.
\newblock {\em Frontiers in computational neuroscience}, 13:432276, 2019.

\bibitem[\protect\citeauthoryear{Takagi and Nishimoto}{2023}]{takagi2023high}
Y.~Takagi and S.~Nishimoto.
\newblock High-resolution image reconstruction with latent diffusion models from human brain activity.
\newblock In {\em CVPR}, 2023.

\bibitem[\protect\citeauthoryear{Tolstikhin \bgroup \em et al.\egroup }{2021}]{tolstikhin2021mlpmixer}
I.~Tolstikhin, N.~Houlsby, A.~Kolesnikov, L.~Beyer, X.~Zhai, T.~Unterthiner, J.~Yung, A.P. Steiner, D.~Keysers, J.~Uszkoreit, M.~Lucic, and A.~Dosovitskiy.
\newblock {MLP}-mixer: An all-{MLP} architecture for vision.
\newblock In {\em NIPS}, 2021.

\bibitem[\protect\citeauthoryear{Urgen and Nizamoğlu}{2022}]{NHA}
B.A. Urgen and H.~Nizamoğlu.
\newblock A large video set of natural human actions for visual and cognitive neuroscience studies and its validation with fmri.
\newblock {\em Brain Sciences}, 13, 2022.

\bibitem[\protect\citeauthoryear{VanRullen and Reddy}{2018}]{Faces}
R.~VanRullen and L.~Reddy.
\newblock Reconstructing faces from fmri patterns using generative adversarial networks.
\newblock In {\em NC Biology}, 2018.

\bibitem[\protect\citeauthoryear{Wang \bgroup \em et al.\egroup }{2024}]{wang2024mindbridge}
S.~Wang, S.~Liu, Z.~Tan, and X.~Wang.
\newblock Mindbridge: A cross-subject brain decoding framework.
\newblock In {\em CVPR}, pages 11333--11342, 2024.

\bibitem[\protect\citeauthoryear{Wen and Shi}{2016}]{DNV}
H.~Wen and J.~Shi.
\newblock Neural encoding and decoding with deep learning for dynamic natural vision.
\newblock {\em Cerebral Cortex}, 28:4136–4160, 2016.

\bibitem[\protect\citeauthoryear{Xia \bgroup \em et al.\egroup }{2024}]{Xia2024UMBRAEUM}
W.~Xia, R.~de~Charette, C.~Oztireli, and J.H. Xue.
\newblock Umbrae: Unified multimodal brain decoding.
\newblock In {\em ECCV}, 2024.

\bibitem[\protect\citeauthoryear{Zeng \bgroup \em et al.\egroup }{2024}]{zeng2024controllable}
B.~Zeng, S.~Li, X.~Liu, S.~Gao, X.~Jiang, X.~Tang, Y.~Hu, J.~Liu, and B.~Zhang.
\newblock Controllable mind visual diffusion model.
\newblock In {\em AAAI}, 2024.

\bibitem[\protect\citeauthoryear{Zhang \bgroup \em et al.\egroup }{2023}]{DBLP:conf/iccv/ZhangRA23}
L.~Zhang, A.~Rao, and M.~Agrawala.
\newblock Adding conditional control to text-to-image diffusion models.
\newblock In {\em ICCV}, pages 3813--3824. {IEEE}, 2023.

\bibitem[\protect\citeauthoryear{Zhou \bgroup \em et al.\egroup }{2024}]{DBLP:conf/iclr/ZhouDWH24}
Q.~Zhou, C.~Du, S.~Wang, and H.~He.
\newblock {CLIP-MUSED:} clip-guided multi-subject visual neural information semantic decoding.
\newblock In {\em ICLR}, 2024.

\end{thebibliography}

\end{document}